\documentclass[11pt]{article}
\usepackage{latexsym}
\usepackage{amsmath}
\usepackage{multirow}
\usepackage{url}

\newcommand{\set}[1]{\{#1\}}
\usepackage{amsmath}
\usepackage{bm}
\usepackage{amsthm}
\usepackage{amssymb}
\usepackage[latin1]{inputenc}
\usepackage{alltt}
\usepackage{verbatim}
\usepackage{graphicx}
\usepackage{xspace}
\usepackage[authoryear]{natbib}
\usepackage{algorithm}
\usepackage[noend]{algorithmic}
\usepackage{datetime}
\usepackage{bm}

\usepackage[subnum]{cases}

\DeclareMathOperator{\softmax}{softmax}

\usepackage{tikz}

\usepackage{enumitem}

\title{\textit{Log-Linear RNNs} :  Towards Recurrent Neural Networks with Flexible Prior Knowledge} %\\[1ex] (Version 1.0) }
\author{Marc Dymetman \qquad Chunyang Xiao \\
Xerox Research Centre Europe, Grenoble, France\\
\tt{\set{marc.dymetman,chunyang.xiao}@xrce.xerox.com}}
\date{November 2016}
\begin{document}
\maketitle
\begin{abstract} 
We introduce \emph{LL-RNNs} (Log-Linear RNNs), an extension of Recurrent Neural Networks that replaces the softmax output layer by a log-linear output layer, of which the softmax is a special case. This conceptually simple move has two main advantages. First, it allows the learner to combat training data sparsity by allowing it to model words (or more generally, output symbols) as complex combinations of attributes without requiring that each combination is directly observed in the training data (as the softmax does). Second, it permits the inclusion of flexible prior knowledge in the form of \emph{a priori} specified modular features, where the neural network component learns to dynamically control the weights of a log-linear distribution exploiting these features.\par

We conduct experiments in the domain of language modelling of French, that exploit morphological prior knowledge and show an important decrease in perplexity relative to a baseline RNN.\par

We provide other motivating iillustrations, and finally argue that the log-linear and the neural-network components contribute complementary strengths to the LL-RNN: the LL aspect allows the model to incorporate  rich prior knowledge, while the NN aspect, according to the ``representation learning'' paradigm,  allows the model to discover novel combination of characteristics. 

\medskip

\noindent\textit{This is an updated version of the e-print arXiv:1607.02467, in particular now including experiments.}
\end{abstract}
% {\scriptsize \tableofcontents}\newpage
\section{Introduction}

Recurrent Neural Networks \citep[Chapter 10]{Goodfellow2016} have recently shown remarkable success in sequential data prediction and have been applied to such NLP tasks as Language Modelling \citep{Mikolov2010}, Machine Translation \citep{sutskever2014sequence,Bahdanau2014}, Parsing \citep{Vinyals2014}, Natural Language Generation \citep{Wen2015} and Dialogue \citep{vinyals2015neural}, to name only a few. Specially popular RNN architectures in these applications have been models able to exploit long-distance correlations, such as LSTMs \citep{hochreiter1997long,Gers2000} and GRUs \citep{Cho2014}, which have led to groundbreaking performances.

RNNs (or more generally, Neural Networks), at the core, are machines that take as input a real vector and output a real vector, through a combination of linear and non-linear operations. 

When working with symbolic data, some conversion from these real vectors from and to discrete values, for instance words in a certain vocabulary, becomes necessary. However most RNNs have taken an oversimplified view of this mapping. In particular, for converting output vectors into distributions over symbolic values, the mapping has mostly been done through a \emph{softmax} operation, which assumes that the RNN is able to compute a real value for each individual member of the vocabulary, and then converts this value into a probability through a direct exponentiation followed by a normalization.

This rather crude ``softmax approach'', which implies that the output vector has the same dimensionality as the vocabulary, has had some serious consequences. 

To focus on only one symptomatic defect of this approach, consider the following. When using words as symbols, even large vocabularies cannot account for all the actual words found either in training or in test, and the models need to resort to a catch-all ``unknown'' symbol \textit{unk}, which provides a poor support for prediction and requires to be supplemented by diverse pre- and post-processing steps \citep{Luong2014,Jean2015}. Even for words inside the vocabulary, unless they have been witnessed many times in the training data, prediction tends to be poor because each word is an ``island'', completely distinct from and without relation to other words, which needs to be predicted individually. 

One partial solution to the above problem consists in changing the granularity by moving from word to character symbols \citep{Sutskever2011,Ling2015}. This has the benefit that the vocabulary becomes much smaller, and that all the characters can be observed many times in the training data. While character-based RNNs have thus some advantages over word-based ones, they also tend to produce non-words and to necessitate longer prediction chains than words, so the jury is still out, with emerging hybrid architectures that attempt to capitalize on both levels \citep{Luong2016c}.

Here, we propose a different approach, which removes the constraint that the dimensionality of the RNN output vector has to be equal to the size of the vocabulary and allows generalization across related words. However, its crucial benefit is that it introduces \emph{\textbf{a principled and powerful way of incorporating prior knowledge inside the models}}.

The approach involves a very direct and natural extension of the softmax, by considering it as a special case of an \emph{conditional exponential family}, a class of models better known as \emph{log-linear} models and widely used in ``pre-NN'' NLP. We argue that this simple extension of the softmax allows the resulting ``log-linear RNN'' to compound the aptitude of log-linear models for \emph{exploiting} prior knowledge and predefined features with the aptitude of RNNs for \emph{discovering} complex new combinations of predictive traits.
\section{Log-Linear RNNs}

\begin{figure*}[t]
\centering{\includegraphics[clip=true, draft=false, trim=0.0cm 0.0cm 13cm 7.2cm, scale=.5]{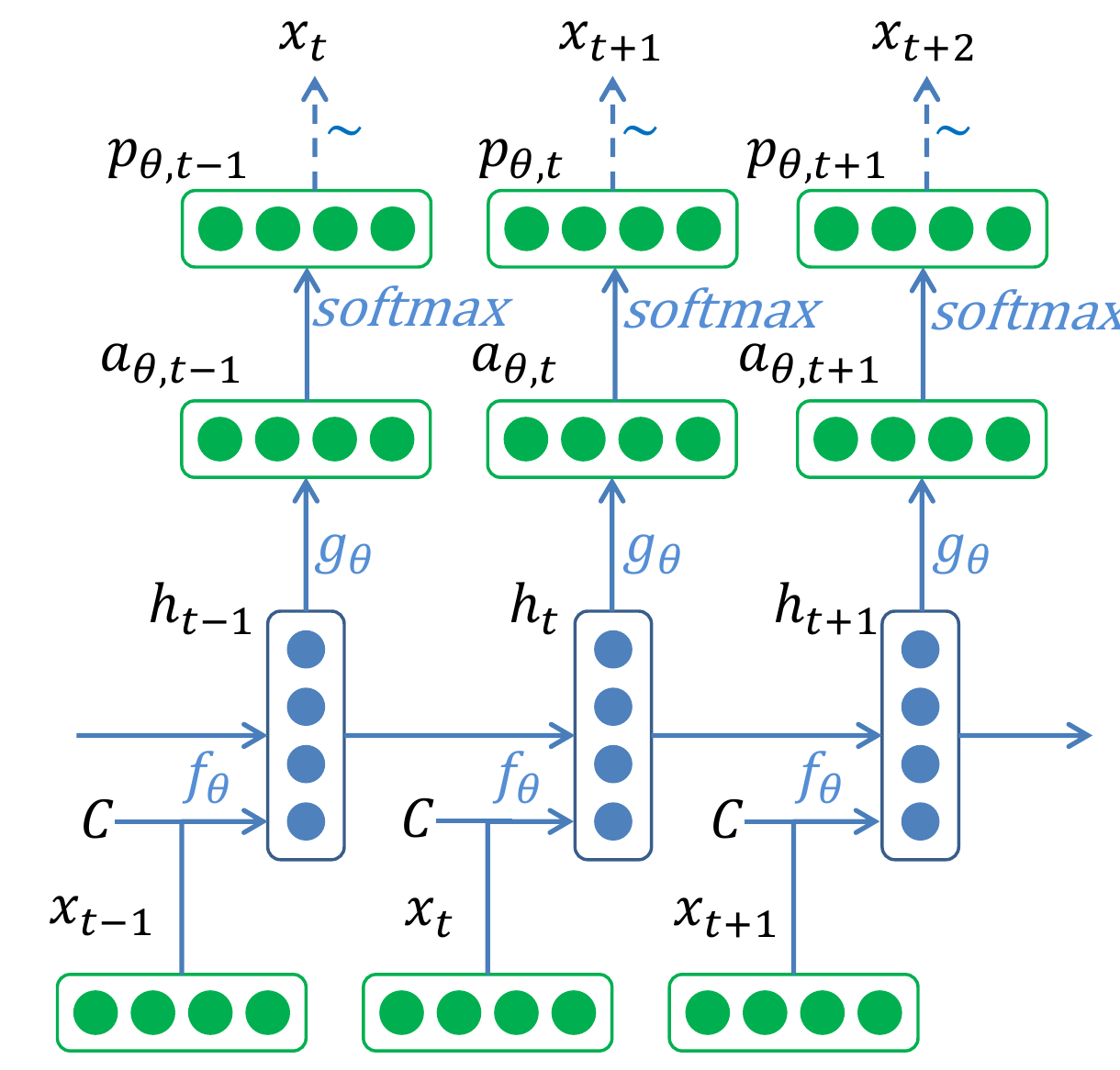}}
\caption{A generic RNN.}
\label{fig:RNN}
\end{figure*}
\subsection{Generic RNNs}
\label{sec:RNN}

Let us first recap briefly the generic notion of RNN, abstracting away from different styles of implementation (LSTM \citep{hochreiter1997long, Graves2012}, GRU \citep{Cho2014}, attention models \citep{Bahdanau2014}, different number of layers, etc.).

An RNN is a generative process for predicting a sequence of symbols $x_1, x_2, \ldots, x_t, \ldots$, where the symbols are taken in some vocabulary $V$, and where the prediction can be conditioned by a certain observed context $C$.
This generative process can be written as:
$$p_\theta(x_{t+1} |C,  x_1, x_2, \ldots, x_t),$$
where $\theta$ is a real-valued parameter vector.\footnote{We will sometimes write this as $p_\theta(x_{t+1} | C; x_1, x_2, \ldots, x_t )$ to stress the difference between the ``context'' $C$ and the prefix $x_1, x_2, \ldots, x_t$. Note that some RNNs are ``non-conditional'', i.e. do not exploit a context $C$.} 
Generically, this conditional probability is computed according to:
\begin{align}
h_t &= f_\theta(C; x_t, h_{t-1}), \label{eq:ht}\\
a_{\theta,t} &= g_\theta(h_t),\label{eq:aTheta}\\
p_{\theta,t} &= \softmax(a_{\theta,t}),\\
x_{t+1} &\sim p_{\theta,t}(\cdot)\,.\label{eq:tilde}
\end{align}

Here $h_{t-1}$ is the hidden state at the previous step $t-1$, $x_t$ is the output symbol produced at that step and $f_\theta$ is a neural-network based function (e.g. a LSTM network) that computes the next hidden state $h_t$ based on $C$, $x_t$, and $h_{t-1}$. The function $g_\theta$,\footnote{We do not distinguish between the parameters for $f$ and for $g$, and write $\theta$ for both.} is then typically computed through an MLP, which returns a real-valued vector $a_{\theta,t}$ of dimension $|V|$. This vector is then normalized into a probability distribution over $V$ through the softmax transformation:
$$\softmax(a_{\theta,t})(x) = 1/Z \exp(a_{\theta,t}(x)),$$
with the normalization factor:
$$Z = \sum_{x'\in V} \exp(a_{\theta,t}(x')),$$
and finally the next symbol $x_{t+1}$ is sampled from this distribution. See Figure~\ref{fig:RNN}.

Training of such a model is typically done through back-propagation of the cross-entropy loss:
$$-\log p_\theta(\bar{x}_{t+1}| x_1, x_2, \ldots, x_t ; C),$$
where $\bar{x}_{t+1}$ is the actual symbol observed in the training set.
\subsection{Log-Linear models}
\subsubsection*{Definition}
Log-linear models play a considerable role in statistics and machine learning; special classes are often known through different names depending on the application domains and on various details: \emph{exponential families} (typically for unconditional versions of the models) \citep{Nielsen2009} \emph{maximum entropy} models \citep{Berger1996,Jaynes1957}, \emph{conditional random fields} \citep{Lafferty2001}, binomial and multinomial \emph{logistic regression} \citep[Chapter 4]{Hastie2001}. These models have been especially popular in NLP, for example in Language Modelling \citep{Rosenfeld1996}, in sequence labelling \citep{Lafferty2001}, in machine translation \citep{Berger1996,Och2002}, to name only a few.

Here we follow the exposition \citep{Jebara2013}, which is useful for its broad applicability, and which defines a \emph{conditional log-linear model} --- which we could also call a \emph{conditional exponential family} --- as a model of the form (in our own notation):
\begin{equation}\label{eq:loglinear}
p(x\,|\,K,a) = \frac{1}{Z(K,a)}\: b(K, x) \exp \left(a^\top \phi(K, x) \right).
\end{equation}
Let us describe the notation:
\begin{itemize}
\item $x$ is a variable in a set $V$, which we will take here to be discrete (i.e. countable), and sometimes finite.\footnote{The model is applicable over continuous (measurable) spaces, but to simplify the exposition we will concentrate on the discrete case, which permits to use sums instead of integrals.}
We will use the terms \emph{domain} or \emph{vocabulary} for this set.
\item $K$ is the \emph{conditioning variable} (also called \emph{condition}).
\item $a$ is a parameter vector in $\mathbb{R}^d$, which (for reasons that will appear later) we will call the \emph{adaptor} vector.\footnote{In the NLP literature, this parameter vector is often denoted by $\lambda$.}
\item $\phi$ is a \emph{feature function} $(K,x) \rightarrow \mathbb{R}^d$; note that we sometimes write $(x;K)$ or $(K;x)$ instead of $(K,x)$ to stress the fact that $K$ is a condition.
\item $b$ is a nonnegative function $(K,x) \rightarrow \mathbb{R}^{+}$; we will call it the \emph{background} function of the model.\footnote{\citet{Jebara2013} calls it the \emph{prior} of the family.}
\item $Z(K,a)$, called the \emph{partition} function, is a normalization factor:
$$ Z(K,a) = \sum_x b(K, x) \exp \left(a^\top \phi(K, x) \right)\, .$$
\end{itemize}
When the context is unambiguous, we will sometimes leave the condition $K$ as well as the parameter vector $a$ implicit, and also simply write $Z$ instead of $Z(K,a)$; thus we will write:
\begin{equation}
p(x) = \frac{1}{Z}\: b(x) \exp \left(a^\top \phi(x) \right),
\end{equation}
or more compactly:
\begin{equation}\label{eq:loglinear-propto}
p(x) \propto b(x) \exp \left(a^\top \phi(x) \right).
\end{equation}
\subsubsection*{The background as a ``prior''}\label{sec:background_as_prior}
If in equation (\ref{eq:loglinear-propto}) the background function is actually a normalized probability distribution over $V$ (that is, $\sum_x b(x) = 1$) and if the parameter vector $a$ is null, then the distribution $p$ is identical to $b$. 

Suppose that we have an initial belief that the parameter vector $a$ should be close to $a_0$, then by reparametrizing equation (\ref{eq:loglinear-propto}) in the form:
\begin{equation}
p(x) \propto b'(x) \exp \left(a'^\top \phi(x) \right),
\end{equation}
with $b'(x) = b(x) \exp (a_0^\top \phi(x))$ and $a' = a - a_0$, then our initial belief is represented by taking $a' = 0$. In other words, we can always assume that our initial belief is represented by the background probability $b'$ along with a null parameter vector $a'=0$. Deviations from this initial belief are then representation by variations of the parameter vector away from $0$ and a simple form of regularization can be obtained by penalizing some $p$-norm $||a'||_p$ of this parameter vector.\footnote{Contrarily to the generality of the presentation by \citet{Jebara2013}, many presentations of log-linear models in the NLP context do not make an explicit reference to $b$, which is then implicitely taken to be uniform. However, the more statistically oriented presentations \citep{Jordan20XX,Nielsen2009} of the strongly related (unconditional) exponential family models do, which makes the mathematics neater and is necessary in presence of non-finite or continuous spaces. One advantage of the explicit introduction of $b$, even for finite spaces, is that it makes it easier to speak about the prior knowledge we have about the overall process.}

\subsubsection*{Gradient of cross-entropy loss}
An important property of log-linear models is that they enjoy an extremely intuitive form for the gradient of their log-likelihood (aka cross-entropy loss).

If $\bar{x}$ is a training instance observed under condition $K$, and if the current model is $p(x | a, K)$ according to equation (\ref{eq:loglinear}), its likelihood loss at $\bar{x}$ is defined as: $- \log L = - \log p(\bar{x} | a, K)$. Then a simple calculation  shows that the gradient $\frac{\partial \log L}{\partial a}$ (also called the ``Fisher score'' at $\bar{x}$) is given by:
\begin{equation}\label{eq:fisher}
\frac{\partial \log L}{\partial a} = \phi(\bar{x}; K) - \sum_{x\in V} p(x|a, K)\, \phi(x; K).
\end{equation}
In other words, the gradient is minus the difference between the model expectation of the feature vector and its actual value at $\bar{x}$.\footnote{More generally, if we have a training set consisting of $N$ pairs of the form $(\bar{x}_n; K_n)$, then the gradient of the log-likelihood for this training set is given by:
\begin{equation*}
\frac{\partial \log L}{\partial a} = \sum_{n=1}^N \left(\phi(\bar{x}_n; K_n) - \sum_{x\in V} p(x|a, K_n)\, \phi(x; K_n)\right).
\end{equation*}
In other words, this gradient is \emph{the difference between the feature vectors at the true labels minus the expected feature vectors under the current distribution}  \citep{Jebara2013}.
}

\subsection{Log-Linear RNNs }
\newcommand\encircle[1]{%
  \tikz[baseline=(X.base)] 
    \node (X) [draw, shape=circle, inner sep=0] {\strut #1};}

\begin{figure*}[t]
\centering{\includegraphics[clip=true, draft=false, trim=0.0cm 0.5cm 0cm 3cm, scale=.5]{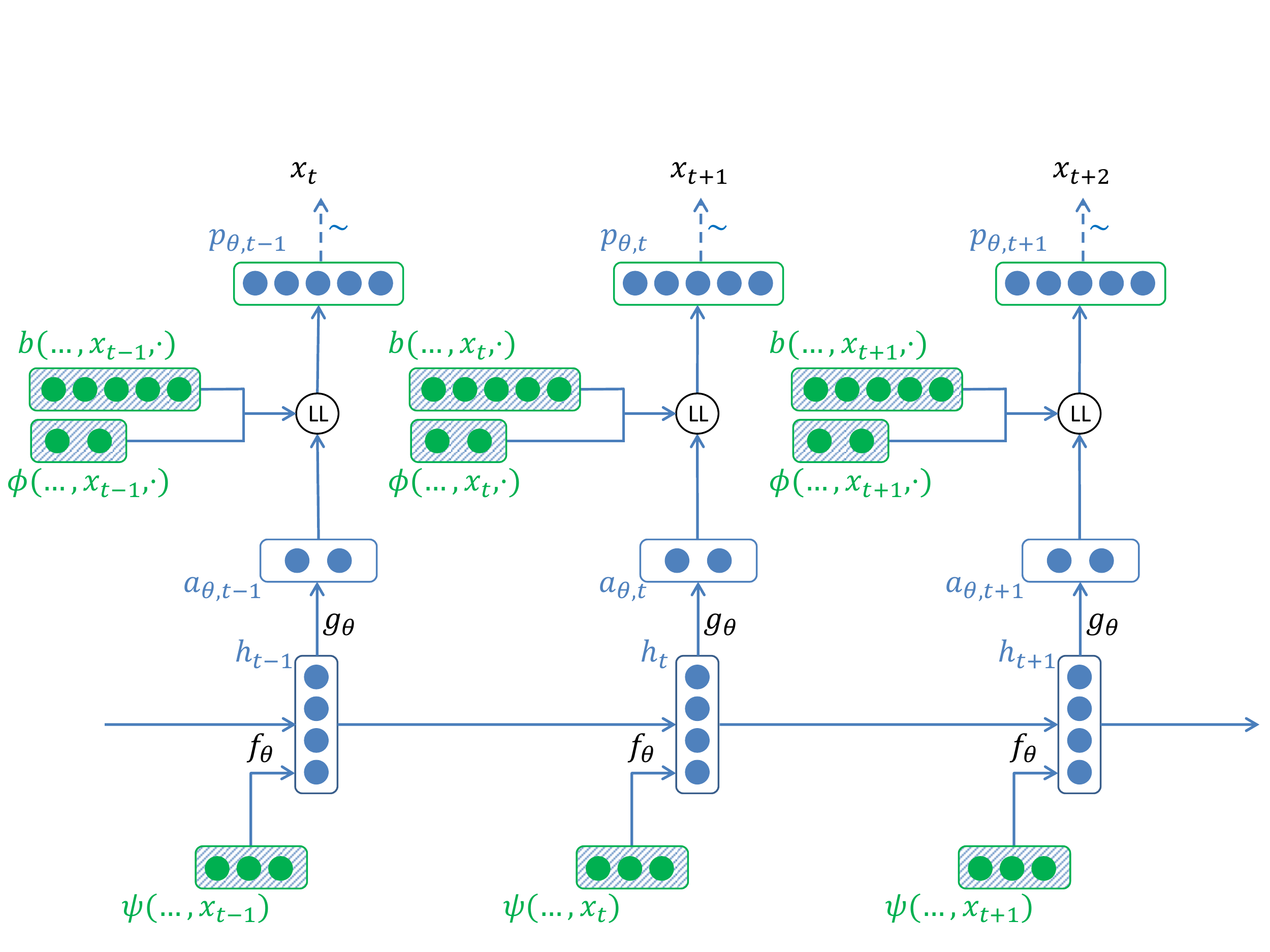}}
\caption{A Log-Linear RNN.}
\label{fig:LL-RNN}
\end{figure*}

We can now define what we mean by a \emph{log-linear RNN}. The model, illustrated in Figure~\ref{fig:LL-RNN}, is similar to a standard RNN up to two differences:

\textbf{The first difference} is that we allow a more general form of input to the network at each time step; namely, instead of allowing only the latest symbol $x_t$ to be used as input, along with the condition $C$, we now allow an arbitrary feature vector $\psi(C, x_1, \ldots, x_t)$ to be used as input; this feature vector is of fixed dimensionality $|\psi|$, and we allow it to be computed in an \emph{arbitrary} (but deterministic) way from the combination of the currently known prefix $x_1, \ldots, x_{t-1}, x_t$ and the context $C$. This is a relatively minor change, but one that usefully expands the expressive power of the network. We will sometimes call the  $\psi$ features the \emph{input features}.

\textbf{The second, major, difference} is the following. We do compute $a_{\theta,t}$ in the same way as previously from $h_t$, however, after this point, rather than applying a softmax to obtain a distribution over $V$, we now apply a log-linear model. While for the standard RNN we had:
$$p_{\theta,t}(x_{t+1}) = \softmax(a_{\theta,t})(x_{t+1}),$$
in the LL-RNN, we define:
\begin{equation}\label{eq:LL-RNN_oneline}
p_{\theta,t}(x_{t+1})  \propto
       b(C, x_1, \ldots, x_t, x_{t+1}) \exp\left({a_{\theta,t}}^\top\, \phi(C, x_1, \ldots, x_t, x_{t+1})\right).
\end{equation}
In other words, we assume that we have \emph{a priori} fixed a certain background function $b(K, x)$, where the condition $K$ is given by $K = (C, x_1, \ldots, x_t )$, and also  defined $M$ features defining a feature vector $\phi(K,x_{t+1})$, of fixed dimensionality $|\phi| = M$. We will sometimes call these features the \emph{output features}.
Note that both the background and the features have access to the context $K = (C, x_1, \ldots, x_t )$. 

In Figure~\ref{fig:LL-RNN}, we have indicated with {\scriptsize \encircle{LL}} (LogLinear) the operation (\ref{eq:LL-RNN_oneline}) that combines $a_{\theta,t}$ with the feature vector $\phi(C, x_1, \ldots, x_t, x_{t+1})$ and the background $ b(C, x_1, \ldots, x_t, x_{t+1})$ to produce the probability distribution $p_{\theta,t}(x_{t+1})$ over $V$. We note that, here, $a_{\theta,t}$ is a vector of size $|\phi|$, which may or may not be equal to the size $|V|$ of the vocabulary, by contrast to the case of the softmax of Figure~\ref{fig:RNN}. 

\par\bigskip\par
\noindent Overall, the LL-RNN is then computed through the following equations:
\begin{align}
h_t &= f_\theta(\psi(C, x_1, \ldots, x_t),  h_{t-1}),\label{eq:LLRNN_line_ftheta}\\
a_{\theta,t} &= g_\theta(h_t),\\
p_{\theta,t}(x) &\propto b(C, x_1, \ldots, x_t, x)\label{eq:LLRNN_line_ptheta}
\hspace*{0.1cm}\cdot\;\exp\left({a _{\theta,t}}^{\top}\ \phi(C, x_1, \ldots, x_t, x)\right),\\
x_{t+1} &\sim p_{\theta,t}(\cdot)\,.
\end{align}

\noindent \textbf{For prediction}, we now use the combined process $p_\theta$, and we train this process, similarly to the RNN case, according to its cross-entropy loss relative to the actually observed symbol $\bar{x}$:
\begin{equation}
-\log p_\theta(\bar{x}_{t+1}| C, x_1, x_2, \ldots, x_t ).
\label{eq:combined_loss}
\end{equation}

\noindent \textbf{At training time}, in order to use this loss for backpropagation in the RNN, we have to be able to compute its gradient relative to the previous layer, namely $a_{\theta,t}$. From equation (\ref{eq:fisher}), we see that this gradient is given by:
\begin{equation}\label{gradient-LL-RNN}
\left(\sum_{x\in V} p(x|a_{\theta,t}, K)\, \phi(K; x)\right) - \phi(K;\bar{x}_{t+1}),
\end{equation}
with $K = C, x_1, x_2, \ldots, x_t$.

This equation provides a particularly intuitive formula for the gradient, namely, as the difference between the expectation of $\phi(K;x)$ according to the log-linear model with parameters $a_{\theta,t}$ and the observed value $\phi(K;\bar{x}_{t+1})$. However, this expectation can be difficult to compute. 
For a finite (and not too large) vocabulary $V$, the simplest approach is to simply evaluate the right-hand side of equation (\ref{eq:LLRNN_line_ptheta}) for each $x\in V$, to normalize by the sum to obtain $p_{\theta,t}(x)$, and to weight each $\phi(K;x)$ accordingly. For standard RNNs (which are special cases of LL-RNNs, see below), this is actually what the simpler approaches to computing the softmax gradient do, but more sophisticated approaches have been proposed, such as employing a ``hierarchical softmax'' \citep{Morin2005}.
In the general case (large or infinite $V$), the expectation term in (\ref{gradient-LL-RNN}) needs to be approximated, and different techniques may be employed, some specific to log-linear models \citep{Elkan2008, Jebara2013}, some more generic, such as contrastive divergence \citep{Hinton2002} or Importance Sampling; a recent introduction to these generic methods is provided in  \citep[Chapter 18]{Goodfellow2016}, but, despite its practical importance, we will not pursue this topic further here.

\begin{comment}
We will come back later to the problem of computing this quantity. \marginpar{\vspace{-4em}
\tiny\sc We will have to say something about how to compute this. Contrastive divergence ? We note that in the oneHot case, the computation is the same as the usual one; for simple adaptor-background processes, there is thus no overhead, apart the issue of computing b(x) for all x's. }
\end{comment}

\subsection{LL-RNNs generalize RNNs}\label{sec:LLRNNs-generalize-RNNs}
 It is easy to see that LL-RNNs generalize RNNs. Consider a finite vocabulary $V$, and the $|V|$-dim ``one-hot'' representation of $x\in V$, relative to a certain fixed ordering of the elements of $V$:
\[
\begin{array}{*6{c}}
\text{oneHot}(x)=[0, &0,&\ldots&\,1,&\ldots&0].\\[-2pt]
&&&\uparrow& &\\[-2pt]
&&&x&&
\end{array}
\]
We assume (as we implicitly did in the discussion of standard RNNs) that $C$ is coded through some fixed-vector and we then define: 
\begin{equation}\label{eq:standard_psi}
\psi(C, x_1, \ldots, x_t) = C \oplus \text{oneHot}(x_t)\:,
\end{equation}
where $\oplus$ denotes vector concatenation; thus we ``forget'' about the initial portion $x_1,\ldots,x_{t-1}$ of the prefix, and only take into account $C$ and $x_t$, encoded in a similar way as in the case of RNNs.

We then define $b(x)$ to be uniformly $1$ for all $x\in V$ (``uniform background''), and $\phi$ to be:
$$\phi(C, x_1, \ldots, x_t, x_{t+1}) = \text{oneHot}(x_{t+1}).$$
Neither $b$ nor $\phi$ depend on $C, x_1, \ldots, x_t$, and we have:
$$p_{\theta,t}(x_{t+1}) \propto b(x_{t+1}) \exp\left({a_{\theta,t}}^\top\, \phi(x_{t+1})\right) = \exp a_{\theta,t}( x_{t+1}),$$
in other words:
$$p_{\theta,t} = \softmax(a_{\theta,t}).$$
Thus, we are back to the definition of RNNs in equations (\ref{eq:ht}-\ref{eq:tilde}). As for the gradient computation of equation (\ref{gradient-LL-RNN}):
\begin{equation}
\left(\sum_{x\in V} p(x|a_{\theta,t}, K)\, \phi(K; x)\right) - \phi(K;\bar{x}_{t+1}),
\end{equation}
it takes the simple form:
\begin{equation}\label{gradient-LL-RNN}
\left(\sum_{x\in V} p_{\theta,t}(x)\, \text{oneHot}(x)\right) - \text{oneHot} (\bar{x}_{t+1}),
\end{equation}
in other words this gradient is the vector $\nabla$ of dimension $|V|$, with coordinates $i\in{1,\ldots,|V|}$ corresponding to the different elements $x_{(i)}$ of $V$, where:
\begin{numcases}
       {\nabla_i =}
       p_{\theta,t}(x_{(i)}) - 1&if $ x_{(i)} = \bar{x}_{t+1}$, \label{numcase:1}\\
       p_{\theta,t}(x_{(i)}) & for the other $x_{(i)}$'s.
\end{numcases}
This corresponds to the computation in the usual softmax case. % \marginpar{\tiny\sc Some reference here ?}

\section{A motivating illustration: rare words}
\label{sec:LLRNN-rare-words}
We now come back to the our starting point in the introduction: the problem of unknown or rare words, and indicate a way to handle this problem with LL-RNNs, which may also help building intuition about these models.

\bigskip

Let us consider some moderately-sized corpus of English sentences, tokenized at the word level, and then consider the vocabulary $V_1$, of size 10K, consisting of the 9999 most frequent words to occur in this corpus plus one special symbol UNK used for tokens not among those words (``unknown words''). 

After replacing the unknown words in the corpus by UNK, we can train a language model for the corpus by training a standard RNN, say of the LSTM type. Note that if translated into a LL-RNN according to section \ref{sec:LLRNNs-generalize-RNNs}, this model has 10K features (9999 features for identity with a specific frequent word, the last one for identity with the symbol UNK), along with a uniform background $b$.

This model however has some serious shortcomings, in particular:
\begin{itemize}
\item Suppose that none of the two tokens \textit{Grenoble} and \textit{37} belong to $V_1$ (i.e. to the 9999 most frequent words of the corpus), then the learnt model cannot distinguish the probability of  the two test sentences: \textit{the cost was 37 euros / the cost was Grenoble euros}.
\item Suppose that several sentences of the form \textit{the cost was NN euros} appear in the corpus, with {NN} taking (say) values {9, 13, 21}, all belonging to $V_1$, and that on the other hand {15} also belongs to $V_1$, but appears in non-cost contexts; then the learnt model cannot  give a reasonable probability to \textit{the cost was 15 euros}, because it is unable to notice the similarity between {15} and the tokens {9, 13, 21}.
\end{itemize}

Let's see how we can improve the situation by moving to a LL-RNN.

We start by extending $V_1$ to a much larger finite set of words $V_2$, in particular one that includes all the words in the union of the training and test corpora,\footnote{We will see later that the restriction that $V$ is finite can be lifted.} and we keep $b$ uniform over $V_2$. 
Concerning the $\psi$ (input) features, for now we keep them at their standard RNN values (namely as in (\ref{eq:standard_psi})).
Concerning the $\phi$ features, we keep the 9999 word-identity features that we had, but not the UNK-identity one; however, we do add some new features (say $\phi_{10000} - \phi_{10020}$):
\begin{itemize}
\item A binary feature $\phi_{10000}(x) = \phi_\text{number}(x)$ that tells us whether the token $x$ can be a number;
\item A binary feature $\phi_{10001}(x) = \phi_\text{location}(x)$ that tells us whether the token $x$ can be a location, such as a city or a country;
\item A few binary features $\phi_\text{noun}(x)$, $\phi_\text{adj}(x)$, ..., covering the main POS's for English tokens. Note that a single word may have simultaneously several such features firing, for instance \textit{flies} is both a noun and a verb.\footnote{Rather than using the notation $\phi_{10000}$, ..., we sometimes use the notation $\phi_\text{number}$, ..., for obvious reasons of clarity.} 
\item Some other features, covering other important classes of words. 
\end{itemize}

Each of the $\phi_1, ..., \phi_{10020}$ features has a corresponding weight that we index in a similar way $a_1, ..., a_{10020}$.

Note again that we \emph{do allow} the features to overlap freely, nothing preventing a word to be both a location and an adjective, for example (e.g. \textit{Nice} in \textit{We visited Nice / Nice flowers were seen everywhere}), and to also appear in the 9999 most frequent words. For exposition reasons (ie in order to simplify the explanations below) we will suppose that a number \textit{N} will always fire the feature $\phi_\text{number}$, but no other feature, apart from the case where it also belongs to $V_1$, in which case it will also fire the word-identity feature that corresponds to it, which we will denote by $\phi_{\tilde N}$, with ${\tilde N} \leq 9999$.

\bigskip

Why is this model superior to the standard RNN one?

To answer this question, let's consider the encoding of \textit{N} in $\phi$ feature space, when $N$ is a number. There are two slightly different cases to look at:
\begin{enumerate}
\item \textit{N} does not belong to $V_1$. Then we have $\phi_\text{10000}=\phi_\text{number}=1$, and $\phi_i=0$ for other $i$'s.
\item \textit{N} belongs to $V_1$. Then we have $\phi_\text{10000}=\phi_\text{number}=1$, $\phi_{\tilde{N}}=1$
and $\phi_i=0$ for other $i$'s.
\end{enumerate}

Let us now consider the behavior of the LL-RNN during training, when at a certain point, let's say after having observed the prefix \textit{the cost was}, it is now coming to the prediction of the next item $x_{t+1} = x$, which we assume is actually a number $\bar{x} =$ \textit{N} in the training sample. 

We start by assuming that \textit{N} does not belong to $V_1$.

Let us consider the current value $a =  a_{\theta,t}$ of the weight vector calculated by the network at this point. According to equation (\ref{eq:fisher}), the gradient is: 
\begin{equation*}
\frac{\partial \log L}{\partial a} = \phi(N) - \sum_x p(x|a)\, \phi(x),
\end{equation*}
where $L$ is the cross-entropy loss and $p$ is the probability distribution associated with the log-linear weights $a$.

In our case the first term  is a vector that is null everywhere but on coordinate $\phi_\text{number}$, on which it is equal to $1$. As for the second term, it can be seen as the model average of the feature vector $\phi(x)$ when $x$ is sampled according to $p(x|a)$.  One can see that this vector has all its coordinates in the interval $[0,1]$, and in fact strictly between $0$ and $1$.\footnote{This last fact is because, for a vector $a$ with finite coordinates, $p(x|a)$ can never be $0$, and also because we are making the mild assumption that for any feature $\phi_i$, there exist $x$ and $x'$ such that $\phi_i(x) =0, \phi_i(x') = 1$; the strict inequalities follow immediately.}
As a consequence, the gradient $\frac{\partial \log L}{\partial a}$ is strictly positive on the coordinate $\phi_\text{number}$ and strictly negative on all the other coordinates. In other words, the backpropagation signal sent to the neural network at this point is that it should modify its parameters $\theta$ in such a way as to increase the $a_\text{number}$ weight, and decrease all the other weights in $a$.

A slightly different situation occurs if we assume now that \textit{N} belongs to $V_1$. In that case $\phi(N)$ is null everywhere but on its two coordinates $\phi_\text{number}$ and $\phi_{\tilde{N}}$, on which it is equal to $1$. By the same reasoning as before we see that the gradient $\frac{\partial \log L}{\partial a}$ is then strictly positive on the two corresponding coordinates, and strictly negative everywhere else. Thus, the signal sent to the network is to modify its parameter towards increasing the $a_\text{number}$ and $a_{\tilde{N}}$ weights, and decrease them everywhere else.

Overall, on each occurrence of a number in the training set, the network is then learning to increase the weights corresponding to the features (either both $a_\text{number}$ and $a_{\tilde{N}}$ or only $a_\text{number}$, depending on whether \textit{N} is in $V_1$ or not) firing on this number, and to decrease the weights for all the other features.  This contrasts with the behavior of the previous RNN model where only in the case of \textit{N} $\in V_1$ did the weight $a_{\tilde{N}}$ change. This means that at the end of training, when predicting the word $x_{t+1}$ that follows the prefix \textit{The cost was}, the LL-RNN network will have a tendency to produce a weight vector $a_{\theta,t}$ with especially high weight on $a_\text{number}$, some positive weights on those $a_{\tilde{N}}$ for which \textit{N} has appeared in similar contexts, and negative weights on features not firing in similar contexts.\footnote{If only numbers appeared in the context \textit{The cost was}, then this would mean \emph{all} ``non-numeric'' features, but such words as \textit{high, expensive, etc.} may of course also appear, and their associated features would also receive positive increments.}

Now, to come back to our initial example, let us compare the situation with the two next-word predictions \textit{The cost was 37} and \textit{The cost was Grenoble}. 
The LL-RNN model predicts the next word $x_{t+1}$ with probability:
$$p_{\theta,t}(x_{t+1}) \propto \exp\left({a_{\theta,t}}^\top\, \phi(x_{t+1})\right).$$
While the prediction $x_{t+1}=\textit{37}$ fires the feature $\phi_\text{number}$, the prediction $x_{t+1}=\textit{Grenoble}$ does not fire any of the features that tend to be active in the context of the prefix \textit{The cost was}, and therefore $p_{\theta,t}(\textit{37}) \gg p_{\theta,t}(\textit{Grenoble})$. This is in stark contrast to the behavior of the original RNN, for which both $37$ and $Grenoble$ were undistinguishable unknown words.

We note that, while the model is able to capitalize on the generic notion of number through its feature $\phi_\text{number}$, it is also able to learn to privilege certain specific numbers belonging to $V_1$ if they tend to appear more frequently in certain contexts. A log-linear model has the important advantage of being able to handle redundant features\footnote{This property of log-linear models was what permitted a fundamental advance in Statistical Machine Translation beyond the initial limited noisy-channel models, by allowing a freer combination of different assesments of translation quality, without having to bother about overlapping assesments \citep{Berger1996,Och2002}.} such as $\phi_\text{number}$ and $\phi_{\tilde{3}}$ which both fire on $3$. Depending on prior expectations about typical texts in the domain being handled, it may then be useful to 
introduce features for distinguishing between different classes of numbers, for instance ``small numbers'' or ``year-like numbers", allowing the LL-RNN to make useful generalizations based on these features. Such features need not be binary, for example a small-number feature could take values decreasing from 1 to 0, with the higher values reserved for the smaller numbers.

\bigskip

While our example focussed on the case of numbers, it is clear that our observations equally apply to other features that we mentioned, such as $\phi_\text{location}(x)$, which can serve to generalize predictions in such contexts as \textit{We are travelling to}. 

In principle, generally speaking, any features that can support generalization, such as features representing semantic classes (e.g. nodes in the Wordnet hierarchy), morphosyntactic classes (lemma, gender, number, etc.) or the like, can be useful.
\section{Some potential applications}
The extension from softmax to log-linear outputs, while formally simple, opens a significant range of potential applications other than the handling of rare words. We now briefly sketch a few directions.
\begin{description}[leftmargin=0cm]
\item [\emph{A priori} constrained sequences] For some applications, sequences to be generated may have to respect certain \emph{a priori} constraints. One such case is the approach to semantic parsing  of \citep{Xiao2016}, where starting from a natural language question an RNN decoder produces a sequential encoding of a logical form, which has to conform to a certain grammar. The model used is implicitely a simple case of LL-RNN,  where (in our present terminology) the output feature vector $\phi$ remains the usual oneHot, but the background $b$ is not uniform anymore, but constrains the generated sequence to conform to the grammar. 

\item [Language model adaptation] We saw earlier that by taking $b$ to be uniform and $\phi$ to be a oneHot, an LL-RNN is just a standard RNN.  The opposite extreme case is obtained by supposing that we already \emph{know} the exact generative process for producing $x_{t+1}$ from the context  $K = C, x_1, x_2, \ldots, x_t$. If we define $b(K; \cdot) = b(K; x)$ to be identical to this true underlying process, then in order to have the best performance in test, it is sufficient for the adaptor vector $a_{\theta,t}$ to be equal to the null vector, because then, according to (\ref{eq:LLRNN_line_ptheta}), $p_{\theta,t}(x) \propto b(K; x)$ is equal to the underlying process. The task for the RNN to learn a $\theta$ such that $a_{\theta,t}$ is null or close to null is an easy one (just take the higher level parameter matrices to be null or close to null), and in this case the adaptor has actually  nothing to adapt to.
\par 
A more interesting, intermediary, case is when $b(K; x)$ is not too far from the true process. For example, $b$ could be a word-based language model (n-gram type, LSTM type, etc.) trained on some large monolingual corpus, while the current focus is on modeling a specific domain for which much less data is available. Then training the RNN-based adaptor $a_\theta$ on the specific domain data would still be able to rely on $b$ for test words not seen in the specific data, but learn to upweight the prediction of words often seen in these specific data.\footnote{For instance, focussing on the simple case of an adaptor over a oneHot $\phi$, as soon as $a_{\theta,t}(K;x)$ is positive on a certain word $x$, then the probability of this word is increased relative to what the background indicates.}

\item [Input features] In a standard RNN, a word $x_t$ is vector-encoded through a one-hot representation both when it is produced as the current output of the network but also when it is used as the next input to the network. 
In section \ref{sec:LLRNN-rare-words}, we saw the interest of defining the ``output'' features $\phi$ to go beyond word-identity features --- i.e. beyond the identification $\phi(x) = \text{oneHot}(x)$ ---, but we kept the ``input'' features as in standard RNNs, namely we kept $\psi(x) = \text{oneHot}(x)$ .
However, let us note an issue there. This usual encoding of the input $x$ means that if $x=37$ has rarely (or not at all) been seen in the training data, then the network will have few clues to distinguish this word from another rarely observed word (for example the adjective \textit{preposterous}) when computing $f_\theta$ in equation \ref{eq:LLRNN_line_ftheta}. The network,  in the context of the prefix \textit{the cost was}, is able to give a reasonable probability to 37 thanks to $\phi$. However,  when assessing the probability of  \textit{euros} in the context of the prefix \textit{the cost was 37}, this is not distinguished by the network from the prefix \textit{the cost was preposterous}, which would not allow \textit{euros} as the next word.
A promising way to solve this problem here is to take $\psi = \phi$, namely to encode the input $x$ using the same features as the output $x$. This allows the network to ``see'' that 37 is a number and that \textit{preposterous} is an adjective, and to compute its hidden state based on this information. We should note, however, that there is no requirement that $\psi$ be equal to $\phi$ in general; the point is that we can include in $\psi$ features which can help the network predict the next word.

\item [Infinite domains] In the example of section \ref{sec:LLRNN-rare-words}, the vocabulary $V_2$ was large, but finite. This is quite artificial, especially if we want to account for words representing numbers, or words taken in some open-ended set, such as entity names. 
Let us go back to the equation (\ref{eq:loglinear}) defining log-linear models, and let us ignore the context $K$ for simplicity:
$p(x|a) = \frac{1}{Z(a)}\: b(x) \exp \left(a^\top \phi(x) \right),$
with $Z(a) = \sum_{x\in V} b(x) \exp \left(a^\top \phi(x) \right)$.
When $V$ is finite, then the normalization factor  $Z(a)$ is also finite, and therefore the probability $p(x|a)$ is well defined; in particular, it is well-defined when $b(x)=1$ uniformly. However, when $V$ is (countably) infinite, then this is unfortunately not true anymore. For instance, with  $b(x)=1$ uniformly, and with $a=0$, then $Z(a)$ is infinite and the probability is undefined. 
By contrast, let's assume that the background function $b$ is in $L_1(V)$, i.e. $\sum_{x\in V} b(x) < \infty$. Let's also suppose that the feature vector $\phi$ is uniformly bounded (that is, all its coordinates $\phi_i$ are such that $\forall x\in V, \phi_i(x) \in [\alpha_i,\beta_i]$, for $\alpha_i,\beta_i \in \mathbb{R}$). Then, for any $a$, $Z(a)$ is finite, and therefore $p(x|a)$ is well-defined. 

Thus, the standard RNNs, which have (implicitely) a uniform background $b$, have no way to handle infinite vocabularies, while LL-RNNs, by using a finite-mass $b$, can. One simple way to ensure that property on tokens representing numbers, for example, is to associate them with a geometric background distribution, decaying fast with their length, and a similar treatment can be done for named entities.

\sloppy
\item [Condition-based priming] Many applications of RNNs, such as machine translation \citep{sutskever2014sequence} or natural language generation \citep{Wen2015}, {etc.}, depend on a condition $C$ (source sentence, semantic representation, etc.). When translated into LL-RNNs, this condition is  taken into account through the input feature vector $\psi(C, x_1, \allowbreak \ldots, x_t) = C  \oplus  \text{oneHot}(x_t)$, see (\ref{eq:standard_psi}), but does not appear in $b(C, x_1, \ldots, , \allowbreak x_t; x_{t+1}) = b(x_{t+1}) = 1$ or $\phi(C, x_1, \ldots, \allowbreak x_t; x_{t+1}) = \text{oneHot}(x_{t+1})$. 
\fussy

However, there is opportunity for exploiting the condition inside $b$ or $\phi$. To sketch a simple example, in NLG, one may be able to predefine some weak unigram language model for the realization that depends on the semantic input $C$, for example by constraining named entities that appear in the realization to have some evidence in the input. Such a language model can be usefully represented through the background process $b(C, x_1, \ldots, x_t; x_{t+1}) = b(C; x_{t+1})$, providing a form of ``priming'' for the combined LL-RNN, helping it to avoid irrelevant tokens.

A similar approach was recently exploited in \cite{Goyal2016}, in the context of a character-based seq2seq LSTM for generating utterances from input ``dialog acts''. In this approach, the background $b$, formulated as a weighted finite-state automaton over characters, is used both for encouraging the system to generate character strings that correspond to possible dictionary words, as well as to allow it to generate strings corresponding to such non-dictionary tokens as named-entities, numbers, addresses, and the like, but only when such strings have evidence in the input dialog act.

%\item [Rich morphology]  should be put in the final discussion

% \item [Combination of experts]  should be put in the final discussion
\end{description}

%%%%%%% Experiments %%%%%%%%%%%%%%%%%%%%%%%%%%%%%%%%%%%%%%%%%%%%%%%%%%%%%%%%%%%
\section{Experiments: French language model using morphological features}
\label{sec:experiments}
\subsection{Datasets}
Our datasets are based on the annotated French corpora\footnote{\url{http://universaldependencies.org/#fr}.}  provided by the \textit{Universal Dependencies} initiative\footnote{\url{http://universaldependencies.org} (Version 1).}. These corpora are tagged at the POS level as well as at the dependency level. In our experiments, we only exploit the POS annotations, and we use lowercased versions of the corpora.

Table \ref{fig:data-stats} shows the sentence sizes of our different datasets, and Table \ref{fig:data-stats} overall statistics in terms of word-tokens and word-types. 

\begin{table}[H]
\centering
\begin{tabular}{ |p{2cm}|p{2cm}|p{1.5cm}|p{1.5cm}|p{1.5cm}|  }
 \hline
		Training	&Validation	&Test1	&Test2\\
 \hline
		13,826  	&728		&298	&1596\\
\hline
\end{tabular}
\caption{Number of sentences in the different datasets. The UD (Universal Dependency) Training set (14554 sents.) is the union of our Training and Validation sets. The UD Validation set is our Test2 set, while the UD Test set is our Test1 set.}
\label{fig:datasets}
\end{table}

\begin{table}[H]
\centering
\begin{tabular}{ |p{2cm}|p{1.5cm}|p{2cm}|p{1.5cm}|p{2cm}|  }
 \hline
Total sentences	&Tokens	 	& Avg. sent. length 	&Types	&Token-type ratio	\\
 \hline
16448	& 463069	 	& 28.15 	& 42894	& 10.80	\\
\hline
\end{tabular}
\caption{Data statistics.}
\label{fig:data-stats}
\end{table}
\subsection{Features}
The corpora provide POS and Morphological tags for each word token in the context of the sentence in which it appears. Table \ref{fig:POS and Morpho features} shows the 52 tags that we use, which we treat as binary features. In addition, we select the $M$ most frequent word types appearing in the entire corpus, and we use $M+1$ additional binary features which identify whether a given word is identical to one of the $M$ most frequent words, or whether it is outside this set. In total, we then use $M+53$ binary features.
 
\begin{table}[t]
\centering
\begin{tabular}{ p{3cm}p{3cm}p{3cm} }
%\hline
POS:ADJ   &Case:Abl      &PronType:Dem   \\
POS:ADP   &Case:Acc  &PronType:In  \\
POS:ADV  &Case:Nom  &PronType:Int  \\
POS:AUX  &Case:Voc  &PronType:Neg  \\
POS:CONJ  &Definite:Def  &PronType:Prs  \\
POS:DET  &Definite:Ind  &PronType:Rel  \\
POS:INTJ  &Degree:Cmp  &Reflex:Yes  \\
POS:NIL  &Gender:Fem  &Tense:Fut  \\
POS:NOUN  &Gender:Masc  &Tense:Imp  \\
POS:NUM  &Gender:Neut  &Tense:Past  \\
POS:PART  &Mood:Cnd  &Tense:Pres  \\
POS:PRON  &Mood:Imp  &VerbForm:Fin  \\
POS:PROPN  &Mood:IndN  &VerbForm:Inf  \\
POS:PUNC  &Mood:SubT  &VerbForm:Part  \\
POS:SCON  &Number:PlurJ &  \\
POS:SYM  &Number:Sing &  \\
POS:VERB  &Person:1 &  \\
POS:X  &Person:2 &  \\
POS:\_  &Person:3 &  \\
%\hline
\end{tabular}
\caption{POS and Morphological features.}
\label{fig:POS and Morpho features}
\end{table}

We collect all the word types appearing in the entire corpus and we associate with each a binary vector of size $M+53$ which is the boolean union of the binary vectors associated with all the tokens for that type. In case of an ambiguous word, the binary vector may have ones on several POS simultaneously.\footnote{Thus for instance, currently, the vector associated with the word \textit{le} has ones not only on the DET (determiner) and the PRON (pronoun) features, but also on the PROPN (proper noun) feature, due to the appearance of the name \textit{Le} in the corpus...}
Thus, here, we basically use the corpus as a proxy for a morphological analyser of French, and we do not use the contextual information provided by the token-level tags.
\subsection{Models}
In these experiments, we use a finite word vocabulary $V$ consisting of the 42894 types found in the entire corpus (including the validation and test sets). We then compare our LL-RNN with a vanilla RNN, both over this vocabulary $V$. Thus none of the models has unknown words.
Both models are implemented in Keras \citep{chollet2015keras} over a Theano \citep{2016arXiv160502688short} backend.\par
The baseline RNN is using one-hot encodings for the words in $V$, and consists of an embedding layer of dimension 256 followed by two LSTM \citep{hochreiter1997long} layers of dimension 256, followed by a dense layer and finally a softmax layer both of dimension $|V|$. The LSTM sequence length for predicting the next word is fixed at 8 words. SGD is done through rmsprop and the learning rate is fixed at 0.001.\par
The LL-RNN has the same architecture and parameters, but for the following differences. First, the direct embedding of the input words is replaced by an embedding of dimension 256 of the representation of the words in the space of the $M+53$ features (that is the input feature vector $\psi$ is of dimension $M+53$). This is followed by the same two LSTMs as before, both of dimension 256. This is now followed by a dense layer of output dimension $M+53$ (the $a_\theta$ weights over the output feature vector $\phi$, here identical to $\psi$.). This layer is then transformed in a deterministic way into a probability distribution over $V$, after incorporation of a fixed background probability distribution $b$ over $V$. This background $b$ has been precomputed as the unigram probability distribution of the word types over the entire corpus.\footnote{We thus use also the test corpora to estimate these unigram probabilities. This is because the background requires to have some estimate of the probability of all the words it may encounter not only in training but also in test. However, this method is only a proxy to a proper estimate of the background for all possible words, which we leave to future development. We note that, similarly, for the baseline RNN, we need to know before hand all words it may encounter, otherwise we have to resort to the UNK category, which we did not want to do in order to be able to do a direct comparison of perplexities.}
\subsection{Results}
Table \ref{table:results} shows the perplexity results we obtain for different cases of the LL-RNNs as compared to the baseline RNN. We use a notation such as LL-RNN (2500) to indicate a value $M=2500$ for the number of frequent word types considered as features. For each model, we stopped the training after the validation loss (not shown) did not improve for three epochs.

\begin{table}[H]
\centering
\begin{tabular}{ |p{3cm}|p{1.5cm}|p{1.5cm}|p{1.5cm}|  }
 \hline
		&Training	&Test1	&Test2\\
\hline
RNN		&4.17  		&6.07	&6.15\\
\hline
LL-RNN (10000)	&4.68 		&5.17	&5.17\\ % LLRNN.26
LL-RNN (5000)	&4.33  		&5.13	&5.12\\
LL-RNN (3000)	&4.65  		&5.11	&5.09\\ % LLRNN.29
LL-RNN (2500)	&4.74  		&5.08	&5.07\\ % LLRNN.28
LL-RNN (2000)	&4.77  		&5.11	&5.10\\ % LLRNN.30
LL-RNN (1000)	&4.84  		&5.13	&5.13\\
LL-RNN (500)	&4.93  		&5.20	&5.18\\ % LLRNN.27
LL-RNN (10)	&5.30  		&5.45	&5.44\\ 
\hline
\end{tabular}
\caption{Log-perplexities per word (base $e$) of different models. The perplexity per word corresponding to a log-perplexity of 5.08 (resp. 6.07)  is 161 (resp. 433).}
\label{table:results}
\end{table}

We observe a considerable improvement of perplexity between the baseline and all the LL-RNN models, the largest one being for $M=2500$ --- where the perplexity is divided by a factor of $433 / 161 \simeq 2.7$ --- with some tendency of the models to degrade when $M$ becomes either very large or very small.

An initial, informal, qualitative look at the sentences generated by the RNN model on the one hand and by the best LL-RNN model on the other hand, seems to indicate a much better ability of the LL-RNN to account for agreement in gender and number at moderate distances (see Table \ref{table:example1}), but a proper evaluation has not yet been performed.

\begin{table}[H]
\begin{small}
\centering
\begin{tabular}{ p{2cm}p{10cm} }
elle        	 &TOPFORM:elle, POS:PRON, POS:PROPN, Gender:Fem, Number:Sing, Person:3, PronType:Prs\\
est         	 &TOPFORM:est, POS:ADJ, POS:AUX, POS:NOUN, POS:PROPN, POS:SCONJ, POS:VERB, POS:X, Gender:Fem, Gender:Masc, Mood:Ind, Number:Sing, Person:3, Tense:Pres, VerbForm:Fin\\
très       	 &TOPFORM:très, POS:ADV\\
souvent     	 &TOPFORM:souvent, POS:ADV\\
représentée	 &TOPFORM:@notTop, POS:VERB, Gender:Fem, Number:Sing, Tense:Past, VerbForm:Part\\
en          	 &TOPFORM:en, POS:ADP, POS:ADV, POS:PRON, Person:3\\
réaction   	 &TOPFORM: réaction, POS:NOUN, Gender:Fem, Number:Sing\\
à          	 &TOPFORM:à, POS:ADP, POS:AUX, POS:NOUN, POS:VERB, Mood:Ind, Number:Sing, Person:3, Tense:Pres, VerbForm:Fin\\
l'          	 &TOPFORM:l', POS:DET, POS:PART, POS:PRON, POS:PROPN, Definite:Def, Gender:Fem, Gender:Masc, Number:Sing, Person:3, PronType:Prs\\
image       	 &TOPFORM:image, POS:NOUN, Gender:Fem, Number:Sing\\
de          	 &TOPFORM:de, POS:ADP, POS:DET, POS:PROPN, POS:X, Definite:Ind, Gender:Fem, Gender:Masc, Number:Plur, Number:Sing, PronType:Dem\\
la          	 &TOPFORM:la, POS:ADV, POS:DET, POS:NOUN, POS:PRON, POS:PROPN, POS:X, Definite:Def, Gender:Fem, Gender:Masc, Number:Sing, Person:3, PronType:Prs\\
république 	 &TOPFORM:république, POS:NOUN, POS:PROPN, Gender:Fem, Number:Sing\\
en          	 &TOPFORM:en, POS:ADP, POS:ADV, POS:PRON, Person:3\\
1999        	 &TOPFORM:1999, POS:NUM\\
.           	 &TOPFORM:., POS:PUNCT\\
\end{tabular}
\end{small}
\caption{Example of a sentence generated by LL-RNN (2500). The right column shows the non-null features for each word. Note that \textit{représentée}, which is not among the most frequent 2500 words (TOPFORM:@notTop), has proper agreement (Gender:Fem, Number:Sing) with the distant pronoun \textit{elle}.\protect\footnotemark}
\label{table:example1}
\end{table}
\footnotetext{As a side remark, we observe some flawed features, due to a small number of gold-annotation errors, such as the fact that \textit{à} appears both with the correct POS:ADP (adposition - a generic term covering prepositions), but also some impossible POS's (AUX, NOUN,VERB). We have not attempted to filter out these (relatively rare) gold-annotation mistakes, but doing so could only improve the results.}

\section{Discussion}

LL-RNNs simply extend RNNs by replacing the softmax parametrization of the output with a log-linear one, but this elementary move has two major consequences. 

The first consequence is that two elements $x, x' \in V$, rather than being individuals without connections, can now share attributes. This is a fundamental property for linguistics, where classical approaches represent words as combination of ``linguistic features'', such as POS, lemma, number, gender, case, tense, aspect, register, etc. With the standard RNN softmax approach, two words that are different on even a single dimension have to be predicted independently, which can only be done effectively in presence of large training sets. In the LL-RNN approach, by associating different $\phi$ features to the different linguistic ``features'', the model can learn to predict a plural number based on observation of plural numbers, an accusative based on the observation of accusatives, and so on, and then predict word forms that are combinations that have never been observed in the training data. We saw an example of this phenomenon in the experiments of section \ref{sec:experiments}.\footnote{Similar observations have been done, in the quite different ``factored'' model recently proposed by \citet{Garcia-Martinez2016}.}
If the linguistic features encompass  semantic classes (possibly provided by Wordnet, or else by semantically-oriented embeddings) then generalizations become possible over these semantic classes also. By contrast, in the softmax case, not only the models are deficient in presence of sparsity of training data for word forms, but they also require to waste capacity of the RNN parameters $\theta$ to make them able to map to the large $a_{\theta}$ vectors that are required to discriminate between the many elements of $V$; with LL-based RNNs, the parametrization $a_{\theta}$ can in principle be smaller, because fewer $\phi$ features need to be specified to obtain word level predictions.

The second consequence is that we can exploit rich prior knowledge through the input features $\psi$, the background $b$, and the output features $\phi$. We already gave some illustrations of incorporating prior knowledge in this way, but there are many other possibilities. For example, in a dialogue application that requires some answer utterances to contain numerical data that can only be obtained by access to a knowledge base, a certain binary ``expert feature'' $\phi_{e}(K;x)$ could take the value 1 if and only if $x$ is \emph{either} a non-number word \emph{or} a specific number $n$ obtained by some (more or less complex) process exploiting the context $K$ in conjunction with the knowledge base. In combination with a background $b$ and other features in $\phi$, who would be responsible for the linguistic quality of the answer utterance, the $\phi_{e}$ feature, when activated, would ensure that \emph{if} a number is produced at this point, it is equal to $n$, but would not try to decide at exactly \emph{which} point a number should be produced (this is better left to the ``language specialists'': $b$ and the other features). \emph{Whether} the feature $\phi_{e}$ is activated would be decided by the RNN: a large value of the $e$ coordinate of $a_{\theta,t}$ would activate the feature, a small (close to null) value deactivate it.\footnote{The idea is reminiscent of the approach of \citet{Le2016}, who use LSTM-based mixtures of experts for a similar purpose; the big difference is that here, instead of using a linear mixture, we use a ``log-linear mixture'', i.e. our features are combined multiplicatively rather than additively, with exponents given by the RNN, that is they are ``collaborating'', while in their approach the experts are ``competing'': their expert corresponding to $\phi_e$ needs to decide on its own at which exact point it should produce the number, rather than relying on the linguistic specialist to do it.\\
This ``multiplicative'' aspect of the LL-RNNs can be related to the \emph{product of experts} introduced by \citet{Hinton2002}. However, in his case, the focus is on learning the individual experts, which are then combined through a direct product, not involving exponentiations, and therefore not in the log-linear class. In our case, the focus is on exploiting predefined experts (or features), but on letting a ``controlling'' RNN decide about their exponents.}
\bigskip

We conclude by a remark concerning the complementarity of the log-linear component and the neural network component in the LL-RNN approach. On its own, as has been amply demonstrated in recent years, a standard softmax-based RNN is already quite powerful. On \emph{its} own, a stand-alone log-linear model is \emph{also} quite powerful, as older research also demonstrated. Roughly, the difference between a log-linear model and a LL-RNN model is that in the first, the log-linear weights (in our notation, $a$) are fixed after training, while in the LL-RNN they dynamically vary under the control of the neural network component.\footnote{Note how a standard log-linear model with oneHot features over $V$ would not make sense: with $a$ fixed, it would always predict the same distribution for the next word. By contrast, a LL-RNN over the same features does make sense: it is a standard RNN. Standard log-linear models have to employ more interesting features.}
However, the strengths of the two classes of models lie in different areas. The log-linear model is very good at exploiting prior knowledge in the form of complex features, but it has no ability to discover new combinations of features. On the other hand, the RNN is very good at discovering which combinations of characteristics of its input are predictive of the output (representation learning), but is ill-equipped for exploiting prior knowledge. We argue that the LL-RNN approach is a way to capitalize on these complementary qualities.

\subsubsection*{Acknowledgments} 
We thank Salah Ait-Mokhtar, Matthias Gall\'{e}, Claire Gardent, \'{E}ric Gaussier, Raghav Goyal and Florent Perronnin for 
discussions at various stages of this research.

\bibliographystyle{apalike}
\bibliography{local}

\end{document}